# HMM-based Writer Identification in Music Score Documents without Staff-Line Removal

[a]Partha Pratim Roy*, [b]Ayan Kumar Bhunia, [c]Umapada Pal

[a]Dept. of CSE, Indian Institute of Technology Roorkee, India, Email: proy.fcs@iitr.ac.in
[b]Dept. of ECE, Institute of Engineering & Management, Kolkata, India, Email: ayanbhunia007@gmail.com
[c]CVPR Unit, Indian Statistical Institute, Kolkata, India, Email: umapada@isical.ac.in
[a]TEL: +91-1332-284816

## Abstract

Writer identification from musical score documents is a challenging task due to its inherent problem of overlapping of musical symbols with staff-lines. Most of the existing works in the literature of writer identification in musical score documents were performed after a pre-processing stage of staff-lines removal. In this paper we propose a novel writer identification framework in musical score documents without removing staff-lines from the documents. In our approach, Hidden Markov Model (HMM) has been used to model the writing style of the writers without removing staff-lines. The sliding window features are extracted from musical score-lines and they are used to build writer specific HMM models. Given a query musical sheet, writer specific confidence for each musical line is returned by each writer specific model using a log-likelihood score. Next, a log-likelihood score in page level is computed by weighted combination of these scores from the corresponding line images of the page. A novel Factor Analysis-based feature selection technique is applied in sliding window features to reduce the noise appearing from staff-lines which proves efficiency in writer identification performance. In our framework we have also proposed a novel score-line detection approach in musical sheet using HMM. The experiment has been performed in CVC-MUSCIMA data set and the results obtained show that the proposed approach is efficient for score-line detection and writer identification without removing staff-lines. To get the idea of computation time of our method, detail analysis of execution time is also provided.

# 1. Introduction

With the rapid progress of mass digitization and transcription in digital libraries, there is a transformation in the ways that people discover information and conduct research. Due to easy availability of rich resources in digital libraries, researchers are developing advanced inquiries to manipulate digital texts and images in different ways which creates numerous research problems in computation analysis involving such rich historical resources (Malik, Roy, Pal & Kimura, 2013). Among these pieces of research work, an interesting application in document image analysis (DIA) field is writer identification which aims to classify the handwritten documents according to the writer.

There exist many archives of historical documents containing music scores where we need to identify the writer. Identification of original writer of a musical score document is a difficult task compared to that of a handwritten text document. It is due to the fact that the number of musical notes/symbols composed by a writer is less in musical sheet than normal handwritten text documents. Music score documents include graphical elements (e.g. staff-lines, musical symbols) and text (e.g. lyrics, etc.) for annotation purpose of musical notations. Generally in a music-score document, notes are written over staff-lines. This document also contains other symbols like Clefs, Accidentals, Time signatures, Dynamics, text etc. (See Fig.1). Since, these musical symbols are overlapped with staff-lines, it is difficult to separate these symbols. There exist some work on writer identification task for such music score document (Marinai, Miotti & Soda, 2010; Fornes, Llados, Sanchez & Bunke, 2009; Bruder, Ignatova, Milewski, 2004). In most of these approaches, these musical documents were passed through some pre-processing techniques such as staff-line removal method which eases the task of writer identification. On the other hand, it is difficult to remove such staff-lines in degraded and curvilinear musical documents. To the best of our knowledge, earlier pieces of research work have not explored yet the writer identification task without removing staff-lines from music documents.

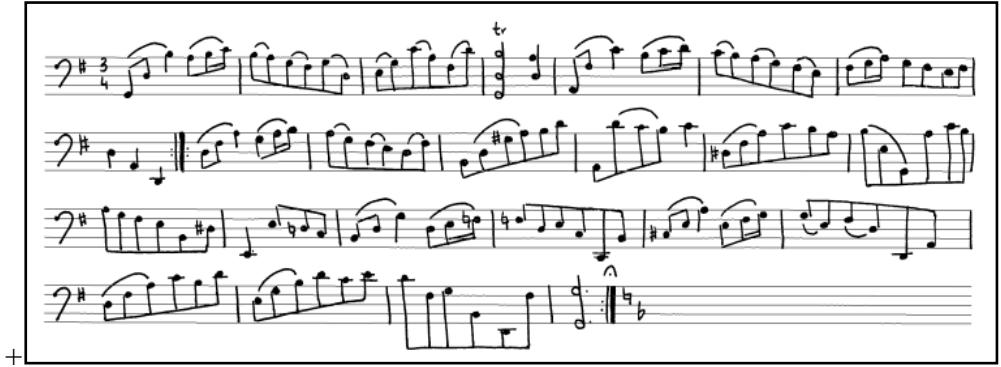
**Fig.1. A musical sheet containing music-symbols overlaid in staff-lines.**

Many pieces of research work exist for writer identification purpose in handwritten text documents (Schomaker & Bulacu, 2004; Schlapbach & Bunke,2007; Schlapbach, Liwicki & Bunke, 2008; Siddiqi & Vincent, 2010; Khan, Tahir, Khelifi, Bouridane, & Almotaeryi, 2017; Wu, Tang, & Bu, 2014). Researchers have developed sophisticated approaches like Markov Random Field (MRF) to remove ruling lines (Cao & Govindaraju, 2007), Hidden Markov Model (HMM) to detect text lines (Bosch, Toselli & Vidal, 2014) in document images. There were also attempts to perform writer identification by avoiding pre-processing techniques such as skewing (Chen & Lopresti, 2013). However, the number of research work towards writer identification in graphical documents is very less. Although the research in graphical documents is primarily focused on Optical Music Recognition (OMR) (Gordo, Fornés, Valveny & Lladós, 2010; Fornés Lladós, Sánchez, Otazu & Bunke, 2010; Fornes, Dutta, Gordo & Llados, 2011), writer identification task in music-score documents opens up new research directions (Bruder, Ignatova, Milewski, 2004; Fornés, Lladós, Sánchez & Bunke, 2008; Marinai, Miotti & Soda, 2010). Writer identification can guide interested musicologists/historians in handling with original drafts, i.e. the original composer of the music. The amount of recent research work handling with removal of staff-lines show the challenging task inherent in writer identification process. The segmentation task is far from satisfactory when the task is for historical manuscripts. Removal of staff-lines in historical manuscripts is not always possible due to the degradation of foreground and background information in musical documents because of ageing. Moreover, preprocessing techniques due to staff-line removal may lose some text/symbol information which will cause the musical symbols appear broken. Thus it needs special care in staff-line removal task for processing musical symbols.

In the past decades, Hidden Markov Models (HMM) has been considered as one of the powerful stochastic approaches. HMM characterizes the temporal observation data that can be discretely or continuously distributed. It has been used successively for modelling sequential data (Gales & Young, 2008; Roy, Bhunia, Das, Dey & Pal, 2016). The efficiency is mostly due to the ability of HMM to cope up with non-linear distortions and incomplete information. Because of such efficiency we have considered HMM in this work. Though HMMs-based techniques have been successfully used in handwriting recognition and writer identification (Schlapbach & Bunke, 2007), it has not been used for writer identification purpose in musical documents.

In this paper, we propose a writer identification framework for music score document which does not require removal of staff-lines from such documents. In our framework, HMM has been used for modelling the writing style of each writer. Given an input musical sheet, the sliding window features are extracted from each musical score-line and then the features are analysed by writer specific HMM models. The log-likelihood scores, returned by writer specific HMM models, are compared and the writer having maximum score is identified. Next, a total log-likelihood score at page level is computed by weighted accumulation of these scores from corresponding line images of that page. In our framework we have also proposed a novel score-line detection approach in musical sheet using HMM. The writer identification performance is improved by incorporating segmentation of lines into portion of lines referred as block-lines in this paper.

To avoid the noise appearing from staff-lines, annotated text, background degradation effect, etc., we consider improving writer identification performance using feature selection. Feature selection step is generally incorporated to eliminate noise and thereby improving the quality of the feature. Some of state-of-the-art feature selection methods include Principal Component Analysis (PCA), Linear Discriminant Analysis (LDA), etc. In (Fischer & Bunke, 2009), PCA technique has been used to improve the cursive handwriting recognition performance. In our framework Factor Analysis-based feature selection technique is applied in sliding window feature to reduce the noise appearing from staff-lines which proved efficiency in writer identification performance.

In this present work we improve our previous conference work (Hati, Roy & Pal, 2014), where some preliminary results were presented, with novel ideas and improved performance of writer identification. The contributions of this extended paper are the following: 1) Recognition without staff-line removal: In many music-script documents (e.g. historical or curvilinear musical staff-lines) it is difficult to segment the staff-line and hence we propose a method where we do not need removal of staff-lines like other existing method. 2) Use of silence zone for better recognition: As silence zone contains only musical staff-lines but no musical-score information hence we propose the use of silence zone in the scheme. 3) Use of block-line segmentation for better accuracy: Instead of using line based writer identification we have used block-line segmentation scheme which improves the writer identification performance. 4) Detection of score and without score zones: For efficient segmentation of music-score lines in document we have used Viterbi forced alignment based decoding algorithm. The score and without score zones of each strip are labeled using Filler model for proper identification of boundaries of score-zones. This helps in proper detection of block-line music score which in-turn improves the writer identification performance. 5) Weighted accumulation of block/line-level score for page level writer identification: The line/block level writer identification scores are combined with a weighted function to find the page level writer identification performance. Except the above major contributions other contributions in this work are: factor analysis based feature selection method, experiment on different synthetic noise added images, experiment on historical music-documents, time complexity analysis using block/line-level writer identification, etc. To the best of our knowledge, such a framework of writer identification in graphical documents has not been used earlier.

The rest of the paper is organized as follows. In Section 2, we describe related work and feature extraction process from music sheet image. The writer identification framework in musical documents using HMM is explained in Section 3. Next, we demonstrate the performance of our framework on CVC-MUSCIMA dataset in Section 4. We also show the performance of our proposed approach in historical musical documents. Finally, conclusions and future work are given.

## 2. Related Work

As mentioned earlier, though there exist few pieces of papers for writer identification of music scores in the literature (Marinai, Miotti & Soda, 2010; Fornés, Lladós, Sánchez & Bunke, 2008; Fornes, Llados, 2010), their performance are poor in a challenging dataset such as the CVC-MUSCIMA dataset (Fornes, Dutta, Gordo & Llados, 2011). To our knowledge, Bruder et al. (Bruder, Ignatova, Milewski, 2004) gave an initial proposal of writer identification from music scores. Features were extracted from music score documents and next the authors used a tree structure for clustering each feature. Finally, K-NN method was used for writer identification purpose. In (Fornés, Lladós, Sánchez, & Bunke, 2008; Fornes, Llados, Sanchez & Bunke, 2009), two different approaches for writer identification were presented. They reported results on a small dataset of 200 images containing 20 different writers (Fornes, Llados, Sanchez & Bunke, 2009). Niitsuma et al. (Niitsuma, Schomaker, van Oosten & Tomita, 2013) presented a work on writer identification in historical musical documents using Contour-Hinge feature space which does not involve explicit score line segmentation. An autoencoder based dimensionality reduction is applied to remove the staff-lines and next contour-hinge features are analysed to identify the writer.

Fornes et al. (Fornés, Lladós, Sánchez, & Bunke, 2008) proposed a writer identification approach after removing staff-lines from music score documents. They achieved 95% accuracy in writer identification while doing experiment on 175 music lines from seven writers. Later, Fornes et al. (Fornes, Llados, Sanchez & Bunke, 2009) proposed a texture feature based approach for writer identification after removing staff-lines. Gabor features and gray-scale co-occurrence matrices features were extracted and K-NN classifier was used for classification. Recently, Gordo et al. (Gordo, Fornés, Valveny, & Lladós, 2010; Gordo, Fornes & Valveny., 2013) proposed an approach based on Bag-of-Notes for writer identification where Blurred Shape Model (BSM) descriptor (Escalera, Fornes, Pujol, Radeva, Sanchez & Llados, 2009) was used to extract feature from each musical symbol. Next, Gaussian Mixture Model based probabilistic codebook was built to represent the musical scores. Support Vector Machine (SVM) was finally used for writer identification purpose.

Texture based feature extraction for music line was introduced in (Fornés, Lladós, Sánchez & Bunke, 2008) where the authors reported 75% accuracy. Using texture feature, 73% identification rate was observed in (Fornes, Llados, Sanchez & Bunke, 2009). Combination of these two approaches showed a better performance with a final score 92% in (Fornés, Lladós, Sánchez, Otazu & Bunke 2010). A more simpler and novel method using shape of music symbol was introduced in (Fornes, Llados, 2010). In (Marinai, Miotti & Soda, 2010), a Self Organizing Maps (SOM) based encoding was used to construct a vocabulary from musical symbols. Next, a cosine similarity was used with a nearest neighbour classifier for writer identification.

ICDAR, 2011 (Fornes, Dutta, Gordo & Llados, 2011) organized a writer identification competition on music score documents after removing staff-lines. In this competition, Hassaıne and Al-Maadeed proposed three different features using edge-based directional probability distribution (Al-Maadeed, Mohammed & Kassis, 2008), grapheme (Al-Ma'adeed, Al-Kurbi, Al-Muslih, Al-Qahtani & Al Kubisi, 2008), and combination of edge and grapheme. An accuracy of 77% was reported by combining both edge and grapheme features. Djeddi et al. (TUA03) proposed a method using nearest neighbour classifier with city block distance metric, SVM, Multilayer Perceptron (MLP) and a classifier combination. They reported 76% accuracy.

Though there exist many methods on writer identification task in musical documents, to the best of our knowledge no work was performed on writer identification task in music documents without removing staff-lines. In this paper we present our work which does not need explicit staff-line detection and their removal. From our experiment we obtained encouraging results.

## 3. Proposed Approach on Writer Identification

In our methodology, first we demonstrate writer identification task in musical documents without removing staff lines. Here, the music page is segmented into individual music score lines containing symbols and staff-lines. Next, sliding-window feature is extracted from each score line to obtain HMM models according to style of each writer. These HMM models are used in testing phase for writer identification task. During feature extraction step, an efficient feature selection process due to Factor Analysis has been used to reduce the noise appearing from staff-lines, annotated text, etc. Finally, writer-specific scores obtained from each score-line is accumulated to get page-level writer-specific score. Later, we propose an HMM-based score line detection from musical scripts which provide better performance than line wise segmentation.

For score line segmentation task, first music pages are passed through noise removal and morphological operations to remove the musical symbols (Fornes, Dutta, Gordo & Llados, 2011). The intensity variation in the resultant music score image provides the staff-line locations which are used in segmenting the score lines. Fig.2 shows segmented score lines obtained from the musical page shown in Fig.1.

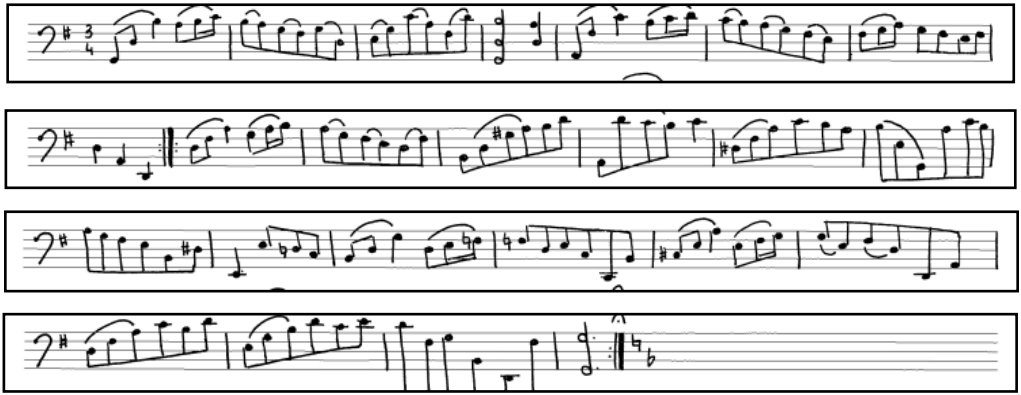

**Fig. 2. Segmentation of score lines from musical sheet shown in Fig. 1. Segmented score lines are marked by rectangular box.**

## 3.1. Writer Identification using Score Line Segmentation

A flowchart of our algorithm for score-line based writer identification is presented in Fig.3. From segmented music-score lines of the document, a sliding window based feature sequence is extracted. Next, the feature sequence is fed to HMM classifier for writer identification purpose. The classification scores obtained from score lines are accumulated with weight assignment to get the final page-level writer identity. These steps are detailed in the following sub-sections.

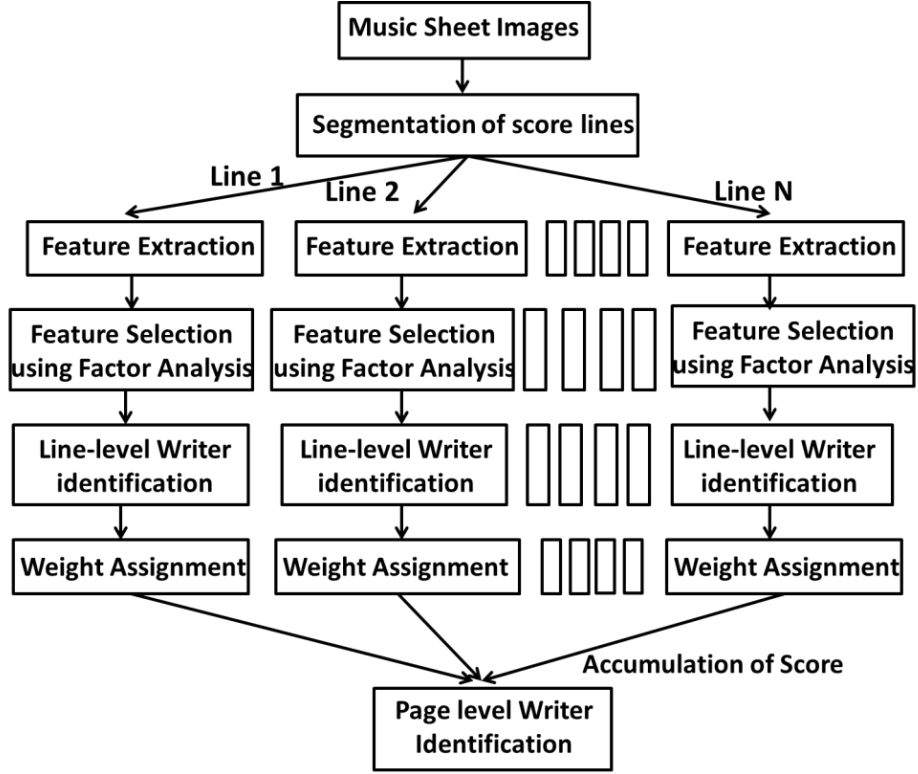

**Fig.3. Block diagram of the writer identification framework**

**3.1.1 Feature extraction from score-lines**

A state-of-the-art sequential feature Local Gradient Histogram (LGH) (Rodríguez-Serrano & Perronnin, 2009) has been used in our framework. LGH feature is popular in handwriting recognition community due to its robustness in recognition. Here, a sliding window moves from left to right direction in an overlapping fashion. Each window is sub-divided into 4 rows and 4 columns in a grid and next a histogram of gradient orientations is calculated from all pixels.

For feature extraction, the horizontal and vertical gradient components $G_x$ and $G_y$ of image $S(x, y)$ are determined as follows.

$$G_x(x,y) = S(x+1, y) - S(x-1, y) \ldots\ldots\ldots (1)$$
$$G_y(x,y) = S(x, y+1) - S(x, y-1) \ldots\ldots\ldots (2)$$

Next, the magnitude *m* and direction $\emptyset$ of gradient are obtained for each pixel with coordinates $(x, y)$ as

$$m(x,y) = \sqrt{G_x^2 + G_y^2} \ldots\ldots\ldots (3) \text{ and } \emptyset(x,y) = tan^{-1}\frac{G_y}{G_x} \ldots\ldots\ldots (4)$$

The field vector $\vec{G} = (G_x, G_y)$, is split into L bin histogram. The histogram is made by summing $m(x, y)$ to the bin indexed by quantized $\emptyset(x, y)$. Thus, the concatenation of the 16 histograms (4 rows x 4 columns) of 8 bins gives a 128-dimensional feature vector.

While extracting the sliding window features, we remove the "silence" zones. Silence zone, in our case contains only musical staff-lines but no musical-score information (See Fig. 4(a)). The sliding-window features in the silence zone do not provide much information specific to a writer. However, it may mislead the writer identification because music-lines from all writers contain silence zones. The silence zones of a score line are marked in Fig. 4(b) with red color. To detect the sliding zones in musical score lines, we check the vertical projection profile of the line image at each sliding window position. Next, if the vertical projection profile for given sliding window position is minimally horizontal, then those sliding window positions are not considered for writer identification purpose. With removal of silence zones the HMM-based framework provided better performance.

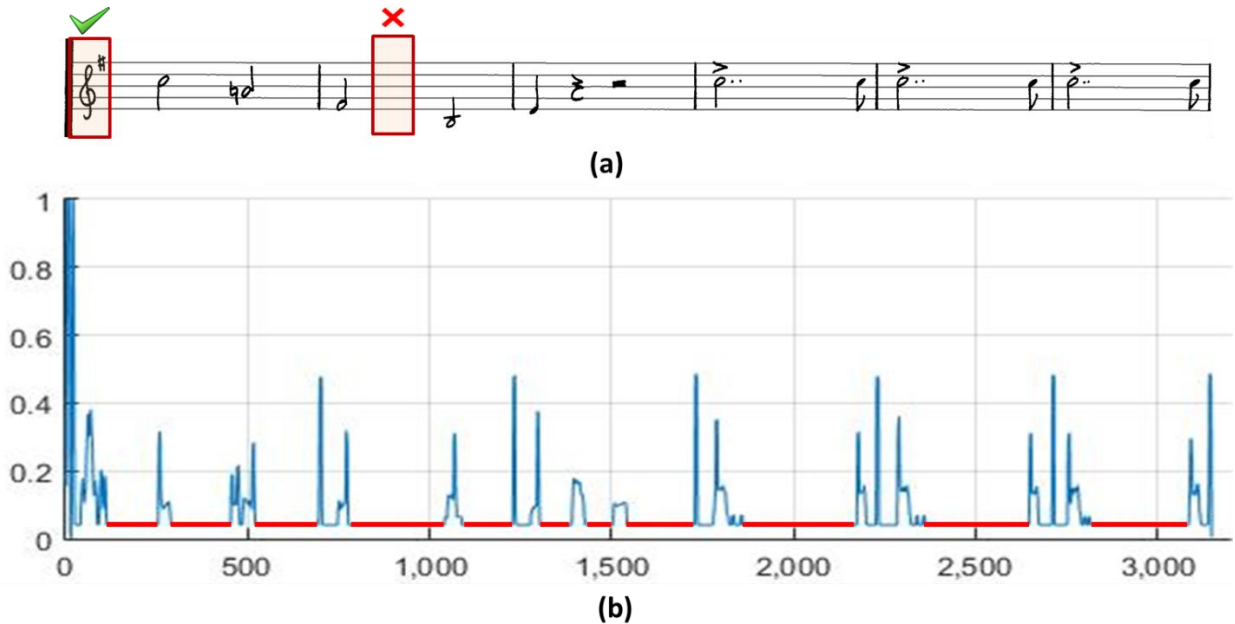

**Fig. 4. (a) Examples show musical-score and silence zones in musical line. (b) Red color marks the silence zones in vertical projection profile of a musical score line.**

### 3.1.2. HMM-based writer identification at line level

Our writer identification framework is built using HMM based classifier where HMM is applied to each segmented music score-lines. An HMM model is created for each writer category. For a classification among *C* categories, we choose the model which best matches the observation from *C* HMMs $\lambda_m = \{A_m, B_m, \pi_m\}$, where $m = 1 \ldots C$, and $\sum_{m=1}^{C} \lambda_m = 1$. Given a test sequence of unknown category, P $(\lambda_i \mid O)$ is calculated for each HMM $\lambda_m$ and select $\lambda_c^*$, where

$$c^* = \arg \max_m P(\lambda_m \mid O) \ldots \ldots \ldots (5)$$

$$P(\lambda_i \mid O) = \frac{P(O \mid \lambda_m) P(\lambda_m)}{P(O)} \ldots \ldots \ldots (6)$$

Where, $P(O)$ is the density function irrespective of the category and is calculated using Eq. (7):

$$P(O) = \sum_{m=1}^{C} P(O \mid \lambda_m) P(\lambda_m) \ldots \ldots \ldots (7)$$

The term $P(O \mid \lambda_m)$ is called the likelihood function for *O* given $\lambda_m$. $P(\lambda_m)$ is called the marginal or prior probability of $\lambda_m$. The Viterbi algorithm provides solution by computing probability $P(O \mid \lambda_m)$ of that sequence generated by $\lambda$. The sequence of LGH feature vector from score line image is used as emission probability of specific writer from HMM states. We used the HTK toolkit (Young, Evermann, Gales, Hain, Kershaw, Liu, Moore, Odell, Ollason, Povey, & Valtchev, 2006) for the HMM implementation. The parameters like, numbers of Gaussian Mixture and state are fixed according to validation dataset.

### 3.1.3. Page level writer identification

From each score-line, the line-level HMM classifier returns a log-likelihood score. Let, *S* = *{S₁, S₂....Sₙ}* be the log-likelihood score of each line image corresponding to N writers. Next, the probability *P* = *{P₁, P....Pₙ}* of the writer's decision is obtained by $P_i = \exp(S_i)$. The writers are ranked as R = *{R₁, R₂....Rₙ}* according to the sorting of probability scores. We show in Fig. 5 the distribution of normalized probability scores (NPS) for four lines of the music document shown in Fig. 1. The normalization of probability score of the writer is obtained by dividing the max score. The NPS scores are shown to better visualize the probability distribution among writers. The groundtruth of writer id of Fig.1 is 9. We noted that HMM estimated correct rank for line 1, 3 and 4 but in case of line 2, some other writers (i.e. 6, 15, 25, and 35) have better rank which is wrong.

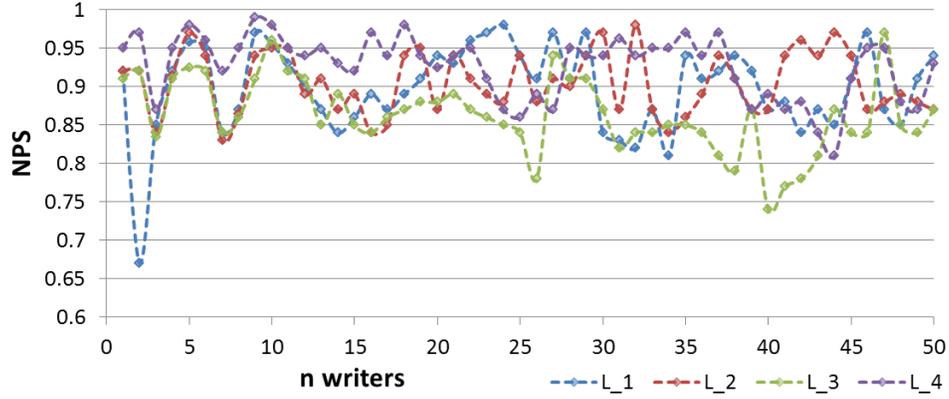

**Fig.5. Normalized probability score distribution for music score line 1 to 4.**

To identify the writer of the music page correctly we combine the line-wise scores in an efficient way. We assign a weight value $W=\{W_{1j}, W_{2j}.....W_{Nj}\}$ to the writers of j$^{th}$ line according to their rank ($R_i$). Finally, a score is obtained for each writer from a page. For a page containing *m* number of score lines, the page-level score $F_i$ of $i^{th}$ writer is estimated using Eq. (8):

$$F_i = \sum_{j=1}^{m}[W_{ij} * P_{ij}] \ldots \ldots \ldots (8)$$

where, $P_{ij}$ and $W_{ij}$ are probability score and weight assignment for j$^{th}$ line, respectively. Different functions of weight assignment were considered in our framework. Table I lists some of these functions. We observed that inverted distance function provides the best result in our framework. In Table I, K is the constant term in uniform function which is nothing but taking simple average of all line scores. N is the number of writer and n is the rank of corresponding writer according to log-likelihood score. In exponential decay function, *a* is the decay constant.

**Table I: Function for weight assignment for page level writer identification**

| Functions | Description |
|---|---|
| **Uniform** | $W = K$ |
| **Inverted distance** | $W = \dfrac{N}{n}$ |
| **Inverted distance squared** | $W = \dfrac{N}{(n^2)}$ |
| **Exponential decay** | $W = \exp(-(a*n))$ |

Finally, the writer having maximum score *max (F₁,F₂....F_N)* is noted and this writer is considered as the target level of that page. Fig. 6 shows the normalized final score of all writers accumulated from Fig. 5. The inverted distance function is used to compute the page level score because of its better performance.

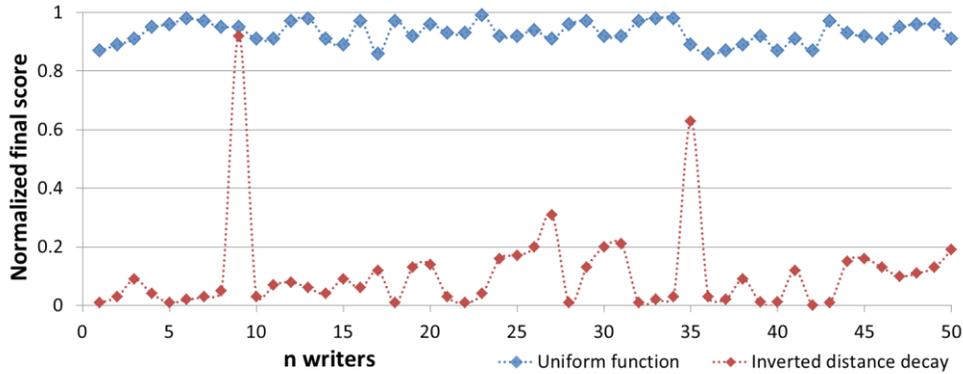

**Fig.6. Normalized final scores for 50 writers obtained from the music page shown in Fig. 1. Blue and Red color denote scores obtained using weighted function and uniform function respectively.**

### 3.2. Feature Selection using Factor Analysis

With the increase of dimension of feature vector in sliding window, it increases the number of unknown parameters of the classifier. Also, the elements of feature vectors are generally correlated due to appearance of similar foreground (staff-lines) in sliding-window feature. To reduce the correlation among features and dimension of feature vector, different feature selection techniques have been introduced earlier. Hence, a feature selection technique like PCA, LDA, etc. is applied to the feature vector sequence obtained from each sliding window. Fig. 7 shows the block diagram of HMM-based writer identification system using feature selection techniques. In this work we found Factor Analysis (FA) suitable because of its better performance. We discuss below briefly the Factor analysis approach for our application.

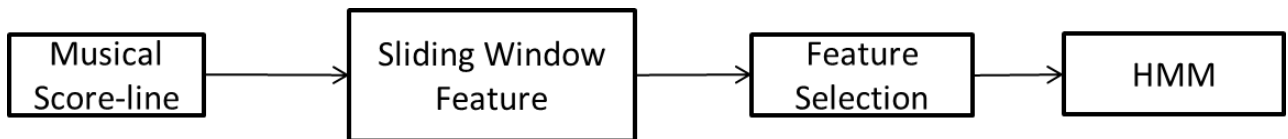

**Fig. 7. Block diagram of the feature selection technique applied in score line.**

Factor Analysis (Hasan & Hansen, 2013) analyzes the relations among a set of random variables of a group. It finds inter-correlations among *n* variables, by postulating a set of common factors. Let $X = \{x_i | i = 1...n\}$ be the feature vector set of n observable variables from the training data with means $\mu_1$, $\mu_2$,.. $\mu_n$. Suppose, for *m* unobserved random variable $y_j$ and for some unknown constants $\lambda_{ij}$, where i = 1 ... n and j = 1 ...m, m < n, we have

$$x_i = \lambda_{i1} y_1 + \lambda_{i2} y_2 + ... + \lambda_{im} y_m + \mu_i + c_i, \ldots \ldots \ldots (9)$$

where $c_i$ is independently noise component with zero mean and finite variance. This noise component c ~ $N(0, \sigma^2 I)$ is assumed to be isotropic and thus the model is equivalent to Probabilistic Principal Component Analysis (PPCA) (Tipping & Bishop, 1999). With factor analysis, a $n \times 1$ feature vector x $\in$ X, the equation can be expressed by Eq. (10),

$$x = Wy + \mu + c \ldots \ldots \ldots (10)$$

where, W is a constant $n \times m$ low rank factor loading matrix that represents $m < n$ bases spanning the subspace with variations in the feature space. $\mu$ is the $n \times 1$ mean vector of x. The latent variable vector y ~ $N(0; I)$ is denoted as writing style factors, that is of size $m \times 1$, i.e., the mean of y is 0, and the covariance is identity matrix. These assumptions help to compute the factor loadings *W*. After estimating the load factors, a Varimax approach is used to change the coordinates which maximize the total sum of the variance of the squared loadings. By this approach, the variance is maximized on the transformed axis. Thus, we find loadings on each factor which is as diverse as possible. One of the advantage of this Factor analysis model is that the writing style factors y, explains the correlation among feature vector x, that can be considered as writer dependent, while the noise component c includes the residual variance of the data and information from staff-lines in musical document.

### 3.3. Musical Score-Line Detection using HMM

The global horizontal projection based line segmentation method constructs histogram by computing sum of all black pixels on every row. Next, individual lines are segmented based on the peak/valley information from the histogram analysis. This projection-based approach will not work on (a) curvy score-lines and (b) degraded situations. To overcome such drawbacks, a piece-wise projection method is used for line segmentation. For this purpose, the musical manuscript is first split into few vertical strips and in each strip the score-lines are detected. We refer the score-lines in each block as block-line. Next in each block-line the score of writer is calculated and finally these are accumulated to find the page-level writer. Note that noise removal technique was not applied in this block-line detection method. In our framework we divide the musical document into vertical strips of equal width. The number of division of strips is decided based on experiment results and it has been found that with 8 divisions we got best accuracy. Fig.8 shows 8 divisions of a musical page and projection profile of 1st division (strip).

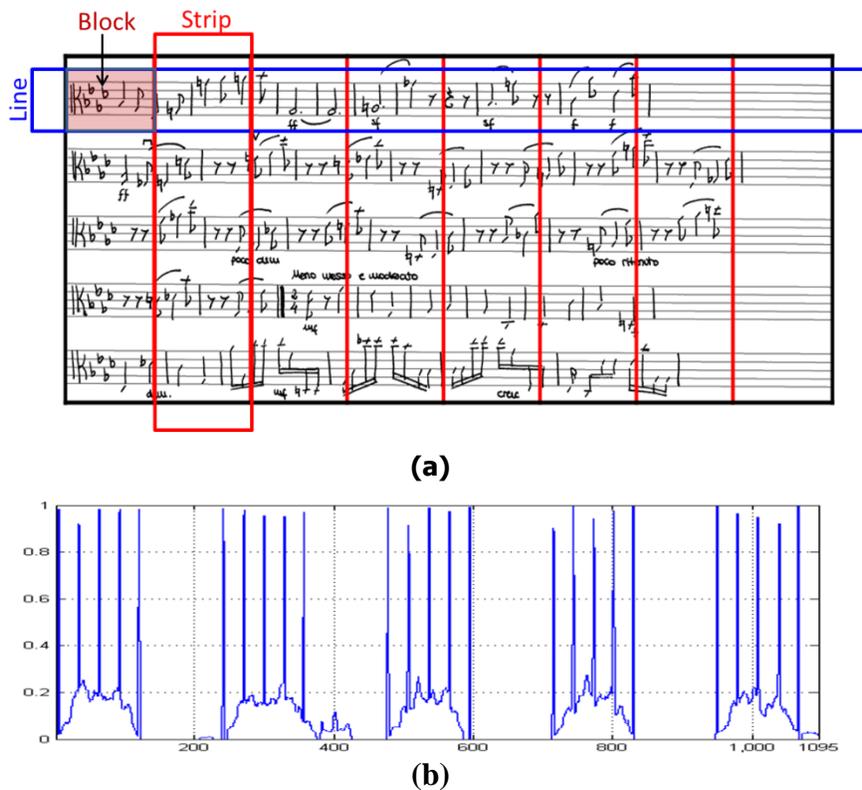

(a)

(b)

**Fig. 8. Image with strip-wise segmentation (a) Musical sheet is divided into 8 strips. The line, block and strip are marked with different colors. (b) Projection profile of 1st strip shows valleys as line-gaps.**

**HMM based music-line detection:** For detecting the boundaries of each music score-line in a page, Viterbi forced alignment based decoding algorithm is used. Viterbi forced alignment is the process of finding the optimal alignment of a set of Hidden Markov Models. For this purpose, we have labelled each strip of musical document into zones of 'Score' and 'Without Score'. The 'Without Score' zone is considered as the empty space between two score lines. For training, first feature vector using LGH is extracted from labelled strip-images. The probability of the 'Score' and 'Without Score' zone models of each strip-image is maximized using Baum-Welch algorithm. Using this information, a filler model is created. The filler model with 'Score' and 'Without Score' zone is shown in Fig. 9(a). Next, Viterbi forced alignment is used in the strip image to align the boundaries of 'Score' and 'Without Score' zone. The segmentation of these zones is refined through iterative alignment and retraining process. We show the segmentation of 'Score' and 'Without Score' zones of a strip image in Fig. 9(b). Blue and yellow colors are used in this figure to demonstrate the 'Score' and 'Without Score' zones. Note that 'Score' zone and block-line refer the same.

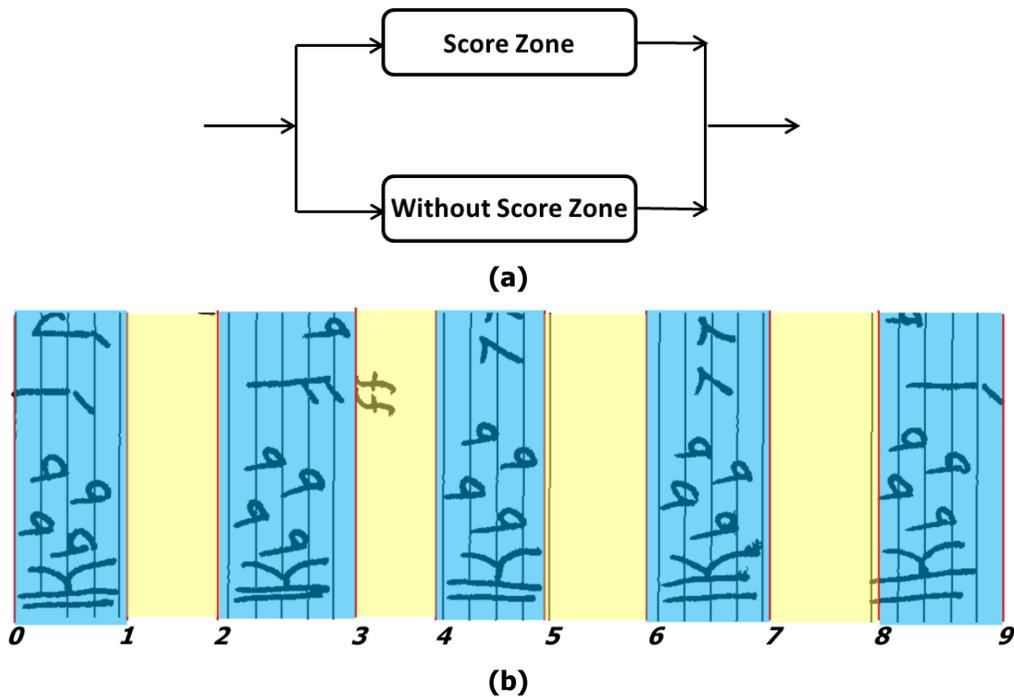

**Fig. 9. (a)Filler model for music block-line detection (b) HMM based alignment of music score lines on 1st strip of Fig. 8. For better readability the strip is rotated in 90° anti-clock wise.**

Next, from each page we have collected all block-lines which are again fed to HMM for block-line level writer identification. After getting the block-line level result we calculated the page level result similarly as discussed above. The steps for writer identification task in musical document image are explained in Algorithm 1.

**Algorithm 1**. Writer Identification in Music Score Documents

---

**Input:** Music sheet images

**Output:** Writer ID

Step 1: Divide a musical sheet into *N* vertical strips.

Step 2: In each strip I = 1 to N

    Step 3: Compute horizontal sliding window feature sequence $W^I_1, W^I_2 ... W^I_c$.

    Step 4: Using feature $W^I_1, W^I_2 ... W^I_c$ HMM-Filler Model is used for Score and Without Score zone classification.

    Step 5: Estimate boundaries of each Score-zone (or block-line) using Viterbi forced alignment.

    Step 6: In each block-line J = 1 to M

        Step 7: Remove silence zones from block-line using projection analysis (discussed in Section 3.1.1).

        Step 8: Compute vertical sliding window feature sequence $F^I_{J1}, F^I_{J2} .. F^I_{JK}$ from remaining portion of block-line.

        Step 9: Apply Factor Analysis in $F^I_{J1}, F^I_{J2} .. F^I_{JK}$ for feature selection and obtain $T^I_{J1}, T^I_{J2} .. T^I_{JK}$

        Step 10: Using feature $T^I_{J1}, T^I_{J2} .. T^I_{JK}$, HMM provides score $S^I_J$ for each writer.

Step 11: Combine all block-line scores $\{S_L\}_{L = 1..MxN}$ of the page using Inverted distance weight to get page level score for each writer.

Step 12: Writer with maximum score is selected as final Writer ID.

---

## 4. RESULT AND DISCUSSION

Our framework of writer identification on musical sheet has been tested on a public "CVC-MUSCIMA" dataset (Gordo, Fornés, Valveny & Lladós, 2010; Fornes, Dutta, Gordo & Llados, 2011) which was used in the ICDAR-2011 and ICDAR-2013 competitions. 50 writers contributed 1,000 music pages in this dataset. 20 different music pages from each writer were considered. The dataset contains two sets of document images, where one set containing documents with staff-lines and other without staff-lines. In our experiment, the experiment is performed in a 10-fold cross-validation mode with 8:1:1 training, validation and testing data. By this, 80% data of dataset was used for training, 10% data for validation and 10% data for testing. In each iteration 800 images (16 pages per writer) were considered for training and 100 images (2 pages per writer) were used for testing. The process is repeated 10 times and final result is obtained by averaging all test data.

As we discussed, the writer identification performance in music-document images are performed in two ways. First, music-score lines are segmented from musical document and next, sliding window features are extracted from these segmented score-lines. No other pre-processing steps like noise removal, staff-line removal, text removal, etc. were applied. The scores obtained from each line-image are combined to get the page-level performance. Second, writer identification performance is obtained by segmenting the page at block-lines. For this purpose, page is divided into number of strips and next, the strips were analyzed for block-line detection. The block-line detection in each strip was performed using HMM. In the following, we detail the different experiment study for writer identification performance. Finally, we show the robustness of the proposed framework for writer identification task in historical musical documents.

**4.1. Writer Identification Performance with Line-level Segmentation**

For writer identification purpose using line-level approach, LGH feature was extracted from each score line image using a sliding window. The overlapping ratio between successive sliding positions was 50%. Next the feature vector sequence from training dataset was used for modeling individual writer style by HMM. Thus, we obtained 50 HMMs corresponding to 50 writers. During testing, feature sequence from a score-line is fed into HMM models and each HMM generates the log-likelihood score. The parameters of sliding window features, like, feature dimension and window width are tuned using validation data. For this purpose, the size of feature dimension in each window position is varied according to angular information. When the orientation is considered as 16 (T=16), we obtain 256 dimensional feature vector. Different dimension of feature size (i.e. 128, 512) were also tested in our experiment (see Fig. 10(a)). Fig. 10(b) presents the results by varying size of the window in validation data. From experiment with validation dataset, we obtained maximum line wise identification accuracy as 60.92% with window size 34 and feature dimension 256. The line-level performance with test data found to be 60.61%. Parameters of HMMs, such as number of states and the number of Gaussian are also validated to govern the proposed architecture (See Fig. 11). From the experiment, we finally decided 64 Gaussians and 4 states for each model.

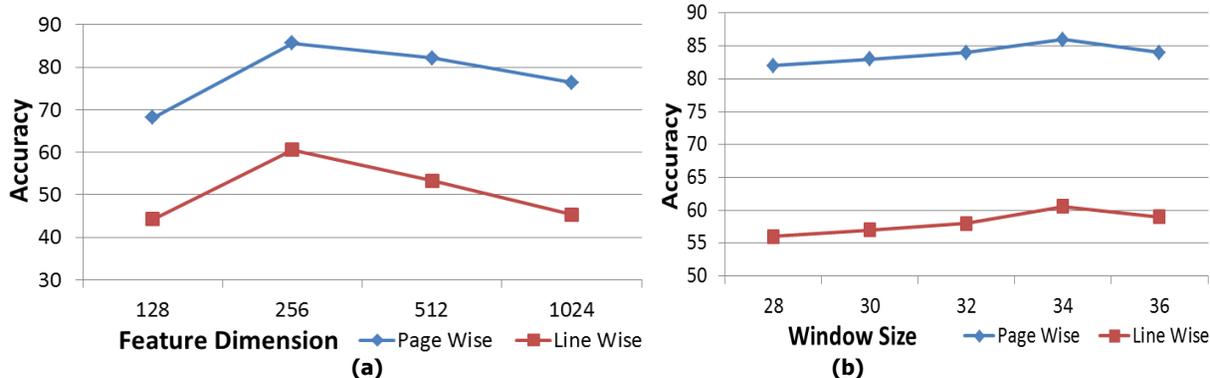
**Fig. 10. Writer identification accuracy against (a) feature dimension and (b) window size of LGH.**

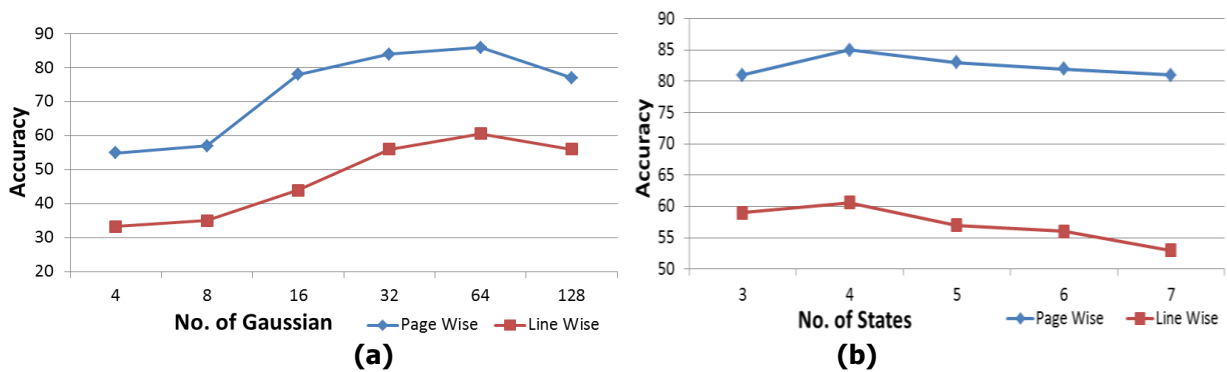
**Fig. 11. Writer identification performance against (a) Gaussians and (b) state numbers.**

### 4.2. Page-wise Writer Identification Performance

Usually, multiple score-lines exist in a music-score document. Some of these score-lines provide better writer-specific information and thus ease the writer identification task. To identify the writer at page-level, line-wise writer identification scores are accumulated. Different weight functions are considered to accumulate the line-based writer scores (see Section 3.1.3). We show the writer identification performance with different weight in Table II. Using 'inverted distance' function, we obtained the best writer identification performance in page-level. A significant improvement in page level was noted (see Fig. 10 and 11) using similar parameter set (i.e. window size, feature dimension, number of states, number of Gaussians) compared to line level.

**Table II: Writer identification performance using different weight functions used in combining line-wise identification scores.**

| Weight function | Accuracy |
|---|---|
| Uniform | 78.96 |
| Inverted distance | 85.67 |
| Inverted distance squared | 84.83 |
| Exponential decay | 83.84 |

A performance analysis with different top choices is shown in Fig. 12. *Top N* indicates that the actual writer is present among the N-best hypotheses. We observed that the page level identification result reached to 100% with 6 top choices, whereas, with this choice number, the line level identification performance was 88.92%.

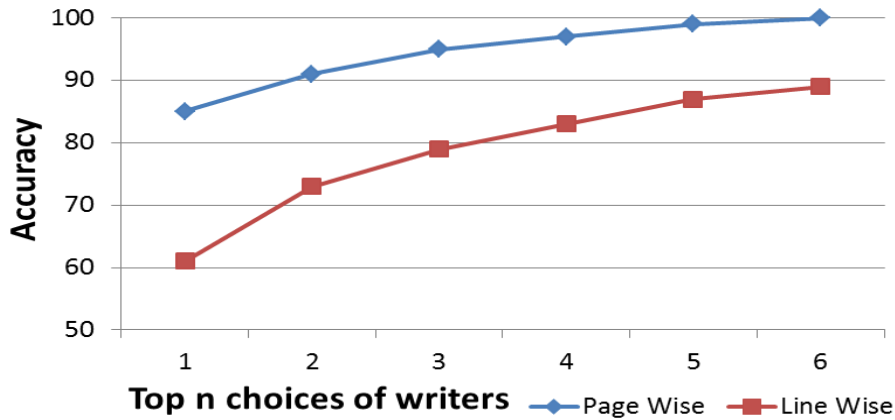

**Fig. 12. Writer identification accuracy with different top choices at Line and Page level.**

### 4.3. Comparison of Factor Analysis with Other Feature Selection Approaches

In our framework, feature selection was used to improve the writer identification performance. To get the maximum variability of the extracted feature we performed feature selection using FA. To compare the performance of the FA models, two feature selection methods, namely PCA and LDA were tested. We found best result using Factor analysis. It provided 86.23% accuracy at page level. Details are shown in Table III.

In Principal Component Analysis (PCA), we are interested to find the directions in which the highest variability in data is observed. This can be done by finding eigenvectors corresponding to the largest eigen values of the covariance matrix. To extract PCA-based features (Cooley & Lohnes, 1971), LGH feature of $n$ dimension from each sliding window patch is considered. Next, $k$ principal components are chosen corresponding to top $k$ (k < n) eigen vectors from $v_j$ by solving the following Eq (11).

$$Av_j = \gamma_j v_j, \quad j = 1, 2, \ldots, n \quad \ldots\ldots\ldots (11)$$

$$A = \frac{1}{p}\sum_{i=1}^{p} z_i z_i^T \quad \ldots\ldots\ldots (12)$$

Where $A$ is the covariance matrix and $z$ represents the normalized vector. The normalization of the input feature vector was performed by unity norm and subtracting by mean. $p$ denotes the number of data points. $\gamma_j$s are the eigen values corresponding to eigen vectors $v_j$. Next, top $k$ principal components are extracted as selected featured for HMM based writer identification. By selecting first few principal components, the dimensionality is reduced while maximum variance is captured. The dimensions with low variance which were not included can be regarded as noise.

In Linear Discriminant Analysis (LDA), the high-dimensional data are projected onto a lower dimensional space by maximizing the separation of data points from different classes while minimizing the dispersion of data from the same class simultaneously. Hence, it achieves maximum class discrimination in the dimensionality-reduced space. The idea of LDA is to find a linear transformation $\theta$ of feature vectors from an n-dimensional space to vectors in an m-dimensional space (m < n) such that the class separability is maximum (Kwon & Narayanan, 2007). The optimization problem is formulated using scatter matrices by following Fisher criterion function:

$$\theta = \underset{\theta}{\arg\max} \frac{\det(\theta^T S_b \theta)}{\det(\theta^T S_W \theta)} \quad \ldots\ldots\ldots (13)$$

$$S_b = \sum_{i=1}^{c} p_i(\mu_i - \mu)(\mu_i - \mu)^T \quad \ldots\ldots\ldots (14)$$

$$S_W = \sum_{i=1}^{c}\sum_{j=1}^{p_i}(x_{ij} - \mu_i)(x_{ij} - \mu_i)^T \quad \ldots\ldots\ldots (15)$$

where, $S_b$ and $S_W$ are the between and within-class scatter matrices, respectively. The columns of the matrix $\theta$ are $S_W$ orthogonal to each other; c is the number of classes. $\mu_i$ and $p_i$ denote the mean and the number of data points in the i[th] class, respectively, n is the dimensionality of the original data space, and μ and p are the global mean and the total number of the data, respectively.

In Table III, a comparative study of these feature selection techniques is shown for writer identification performance at line and page levels. Though the performance using PCA and LDA are similar at line level, LDA showed better performance at page-level writer identification performance. FA based feature selection approach provided the best result at page level performance.

**Table III: Accuracy of different weight functions for combining line-wise identification scores using feature selection techniques.**

|  | **Line level** | **Page level** |
|---|---|---|
| **Baseline Method (HMM with LGH feature)** | 60.61% | 85.67% |
| **Baseline +FA** | **61.68%** | **86.23%** |
| **Baseline +PCA** | 59.72% | 84.65% |
| **Baseline +LDA** | 59.85% | 86.04% |

**4.4. Writer Identification Performance with Block-line level Segmentation**

After performing the HMM-based alignment of the score line on each strip of an image at page level all block-line portions are passed for feature extraction. Fig. 13(a) shows the alignment performance according to number of strips used for dividing the musical sheet. Then LGH feature is extracted from each block-line images using a sliding window with 50% overlapping ratio in successive position. The features are processed and 50 HMMs were obtained according to number of writers. During testing, feature sequence is extracted from a query sequence and HMM provides the log-likelihood score corresponding to each writer. The writer identification performance according to different number of strips are shown in Fig.13(b). We obtained best accuracy with 8 strips. The alignment performance with 8 divisions is 98.48%. We obtained 76.98% accuracy in writer identification at page level with such division.

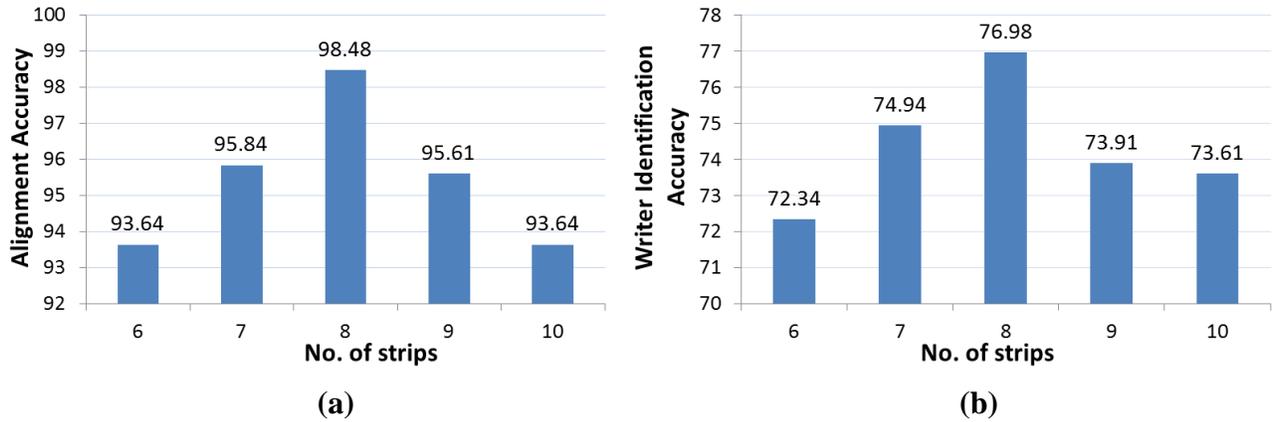

**Fig. 13. (a) HMM-based alignment performance against number of strip-division (b) Full page level writer identification performance against number of strip-division.**

The performance of writer identification against Gaussian number and varying state number are shown in Fig.14. The best results in block-line level and page-level are noted as 30.6% and 76.98% respectively with 256 Gaussian number and 4 states. Fig.15 shows the comparative studies of feature reduction approaches with varying feature dimension. It is to be noted that FA based approach outperforms PCA and LDA approaches. With 256 dimensions in FA we got the best results. The improvement of the result is due to the efficiency in elimination of noise such as staff-lines. Fig.16 demonstrates the performance with silence zone removal process in our framework. The removal of silence zones provided a significant improvement (see Fig. 16(b)) in writer identification in page-level. Finally Fig.17 shows the performance with Top 5 choices in writer identification. Please note that with Top 5 choices the accuracy has been reached to 100% with our HMM-based score-line detection approach, whereas with line-level we reached 100% accuracy with Top 6 choices.

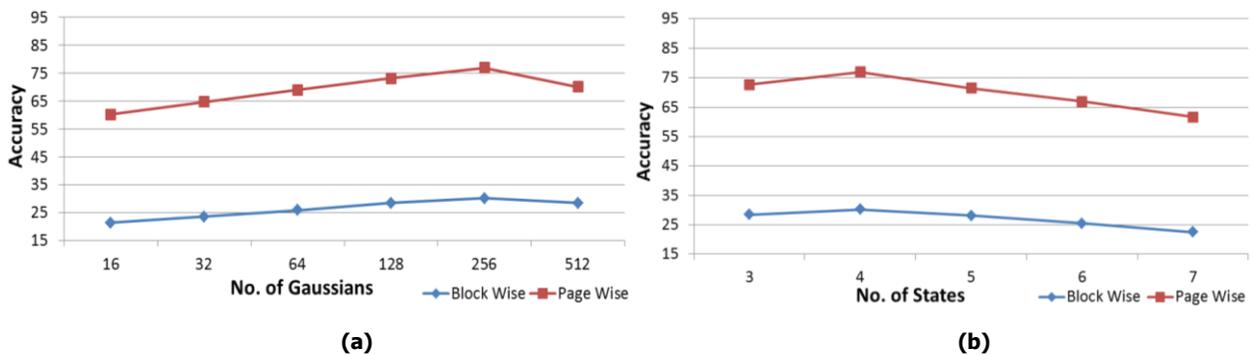

**Fig.14. Writer identification accuracy with different (a) Gaussian and (b) State.**

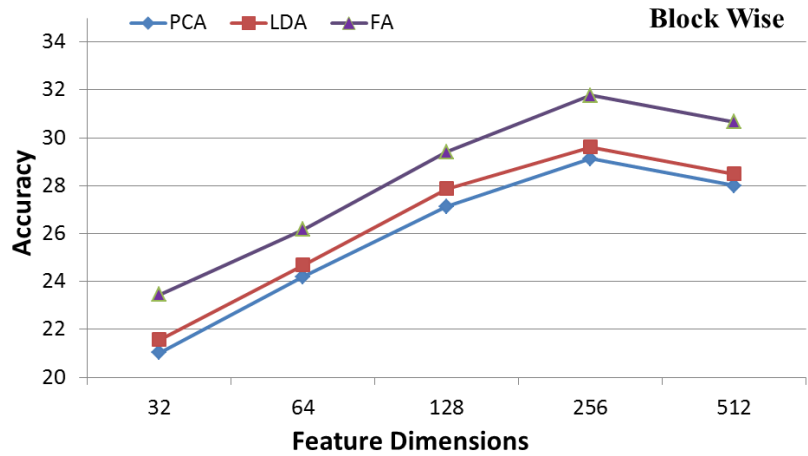

(a)

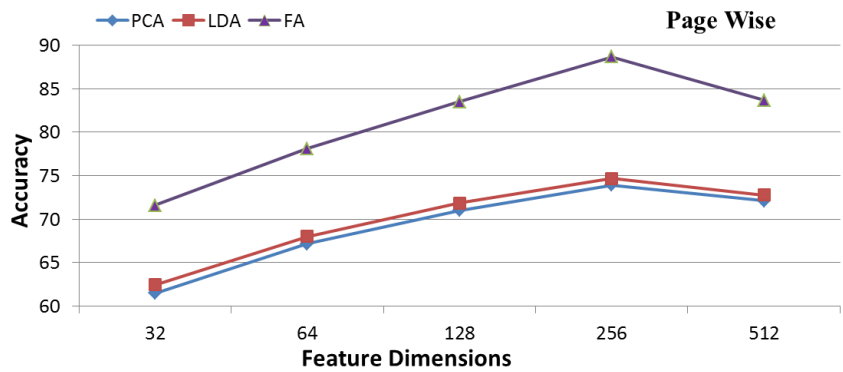

(b)

**Fig.15.** Writer identification performance at (a) block-line and (b) page level against feature dimensions using different feature reduction techniques.

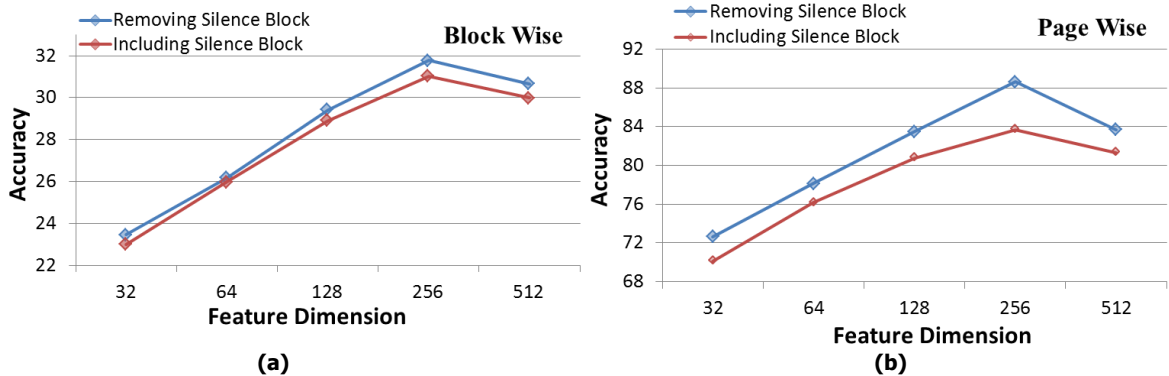

**Fig.16.** Writer identification performances with removal of silence regions at (a) block-line and (b) page levels are shown. FA based feature selection have been used in this experiment.

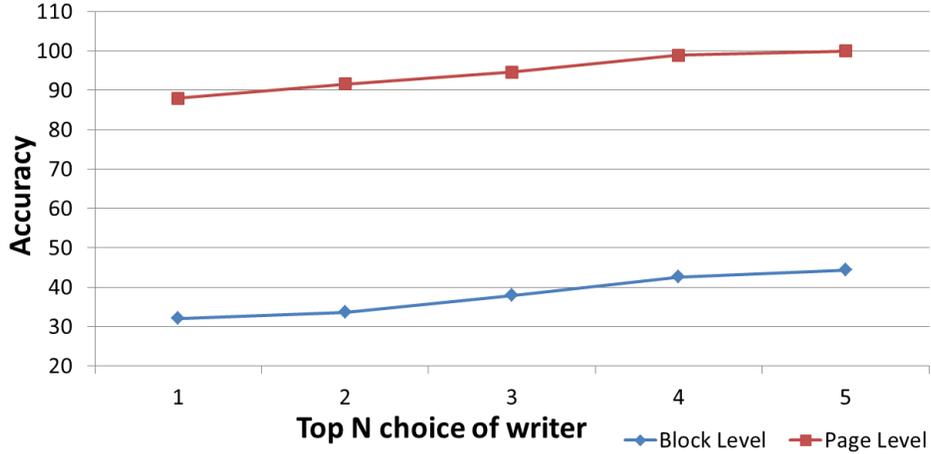

Fig. 17. Writer identification performance with Top 5 choices.

### 4.5. Comparative Analysis

Our HMM based writer identification framework has been compared with different features and classification techniques available in the literature. As mentioned earlier we have used an off-the-shelf feature descriptor (LGH) to evaluate our writer identification performance. LGH has been applied successfully in text recognition in different scripts. From our experiment LGH provided us encouraging results compared to other features. To have an idea we have tested the writer identification performance in musical document using Gabor features. It would be noted that Gabor feature did not provide good results. Also, as an alternative identification framework we have performed our experiments using Gaussian Mixture Models based writer modelling. Details of these methods and performance analysis are discussed in this section.

**Gabor Features:** The GABOR features has been applied successfully in character and text recognition (Chen, Cao, Prasad, Bhardwaj & Natarajan, 2010). The GABOR features is the product of a sinusoid and a Gaussian:

$$g(x, y; \lambda, \sigma, \theta, \varphi, \gamma) = \exp(-\frac{{x'}^2 + \gamma^2 {y'}^2}{2\sigma^2}) \cos(2\pi \frac{x'}{\lambda} + \varphi) \quad \ldots \ldots \ldots (16)$$

Where $x' = x\cos\theta + y\sin\theta, y' = -x\sin\theta + y\cos\theta$. Gabor filter is used in normalized image at four orientations (0°, 45°, 90°and 135°) and then we used the magnitude as the response for feature extraction. After filtering, the image frame is divided equally into 12 rows. Next, we concatenate the features in each grid to have 48 dimensional Gabor features (Chen, Cao, Prasad, Bhardwaj &

Natarajan, 2010). Using Gabor feature we obtained maximum accuracy of 78.45% at page level (using block-line). During this experiment, 256 Gaussian mixtures were used in HMM and FA was used for feature selection (see Fig. 18). The writer identification accuracy without FA was 72.16%.

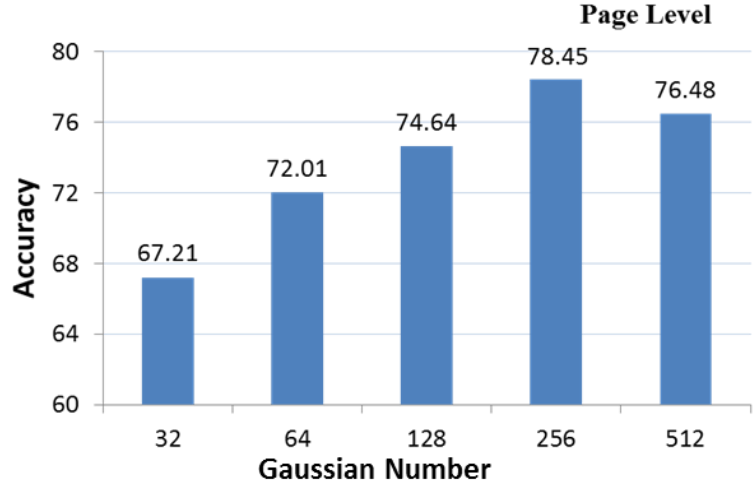

**Fig.18. Writer identification performance with Gabor feature using different Gaussian numbers.**

**GMM-based recognition performance:** Our HMM-based framework is compared with Gaussian Mixture Model (GMM) based approach. Similar to our HMM-based writer models, GMM (Schlapbach & Bunke, 2006) based writer model can also be created for each writer. Sliding window feature extracted from music-score lines are modelled by Gaussian mixture density. The mixture density from a D-dimensional feature vector x, for a writer can be defined by following equation

$$p(\mathbf{x}/\lambda)= \sum_{i=1}^{M} w_i \, p_i(\mathbf{x}). \ldots \ldots \ldots (17)$$

The density is a weighted combination of *M* uni-modal Gaussian densities, $p_i(\mathbf{x})$, each parameterized by a $D \times 1$ mean vector, $\mu_i$, and a $D \times D$ covariance matrix, $C_i$. The parameters of a GMM model are presented as $\lambda = \{w_i, \mu_i, C_i\}, i = 1,...,M$ where $w_i$ are the mixture weights. During decoding, the feature vectors are considered to be independent. Given a sequence of sliding feature vector *X*, the log-likelihood of a model $\lambda$ is defined as

$$\log p(X/\lambda)= \sum_{t=1}^{T} \log p(\mathbf{x}_t|\lambda) \ldots \ldots \ldots (18)$$

During GMM-based writer identification experiment, same parameter setup of HMM were used. Sliding window features using LGH are extracted and these are fed to the GMM. Line-level writer identification scores are next accumulated to provide page level accuracy. Less than 50% accuracy was obtained using GMM framework.

We have measured the scalability of the proposed framework according to number of writers. We noted that HMM-based approach works well for lesser number of writers (see Fig. 19) and we obtained 100% accuracy upto 7 writers in page-level. The line level accuracy was more than 80% with same number of writers. Unlike HMM, the performance falls down in GMM with increasing number of writers. A significant decrease in performance was observed while testing with more than 10 writers. One of the advantages of GMMs over HMMs is its shorter training time. Also, GMMs are less complex, as it consists of only one state and one output distribution function.

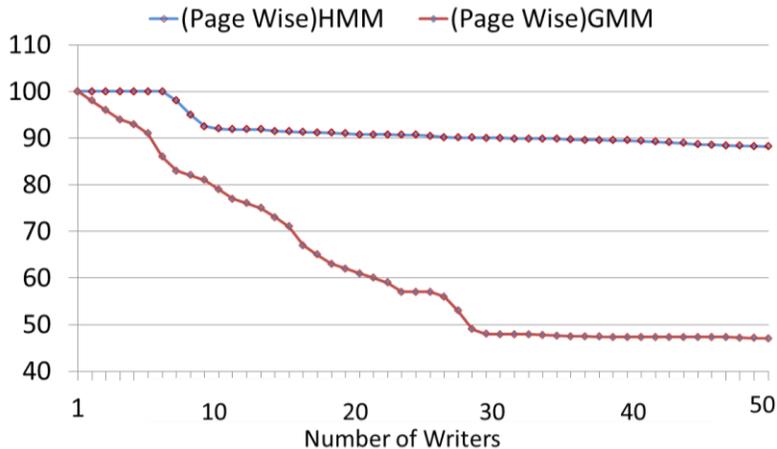

**Fig. 19. Comparison of scalability analysis by increasing number of writers in HMM and GMM framework.**

### 4.6. Error Analysis

Though our system provides a good accuracy overall, there are few situations where identification process is susceptible to error. Sometimes, due to presence of staff-lines and less number of music symbol present in a music-line, it may fail extracting necessary information and hence confusion may appear in some of the music pages. If the difference between writing styles are not significant, the system return wrong classification. One of such example is shown in Fig.20, where due to similar writing style between writer 17 and 42, our system failed. There are some music-score lines which could not be detected properly with HMM-based score-line detection approach. In Fig.21, due to narrow space between two score lines the score-gap detection was not successful.

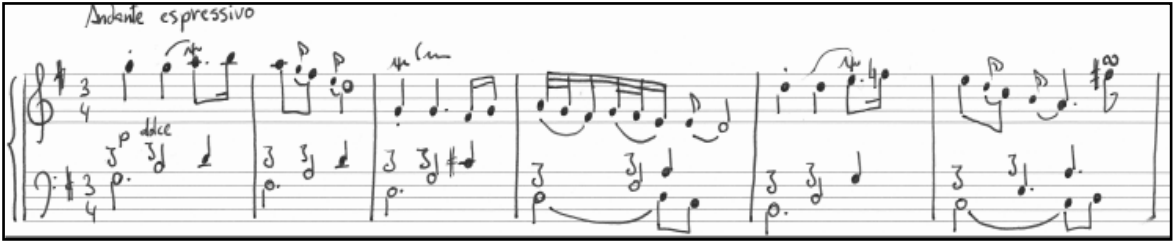

(a)

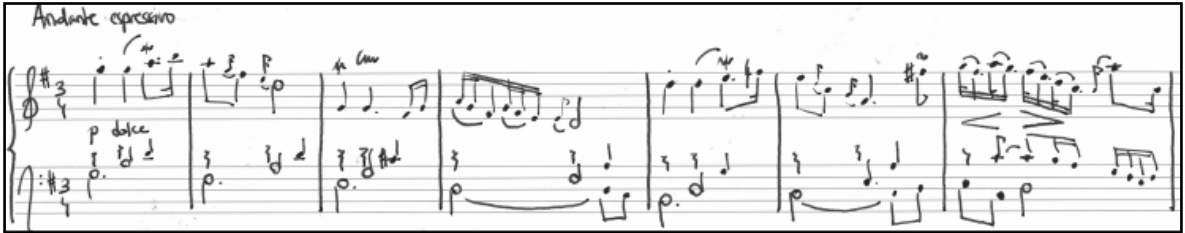

(b)

**Fig. 20. Confusion of writer in images (a) from writer 17 and (b) from writer 42.**

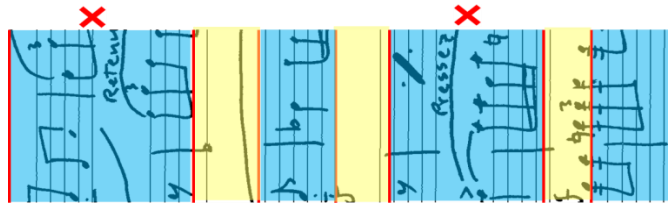

**Fig. 21. Examples of wrong alignment in HMM. The wrong detection of block-lines is marked with 'X' in top of the image.**

To identify the writers which are more prone towards wrong identification (Fig 22), we have used the following formula to measure.

$$Error(\%) = 100 \: X \: \frac{(E-O)}{E} \: ...\:...\:... (19)$$

Where, E (expected accuracy) is the accuracy corresponding to successful identification of all writers (ideal case). O (observed accuracy) is the page wise identification. An error analysis of 50 writers at page-level (using block-line) is shown in Fig. 22. We observed that those writers 14, 27, 29 and 41 are having maximum error rate. Overall error percentage (Error) in page level (using block-line) is 11.35%. The confusion matrix of 50 writers is shown in Fig.23. Color chart demonstrate the writer-wise identification performance in the dataset.

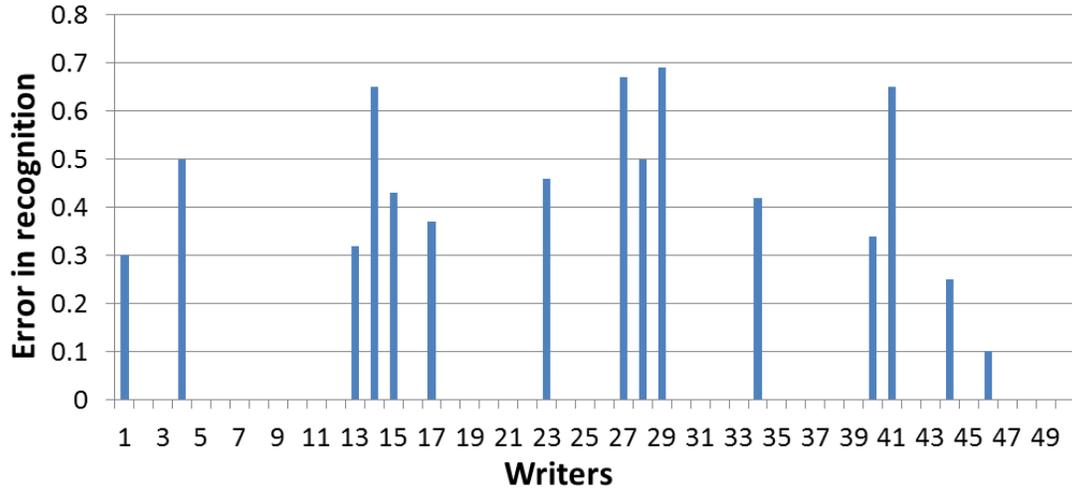

**Fig. 22. Error in page wise identification (using block-line segmentation approach) for different writers**

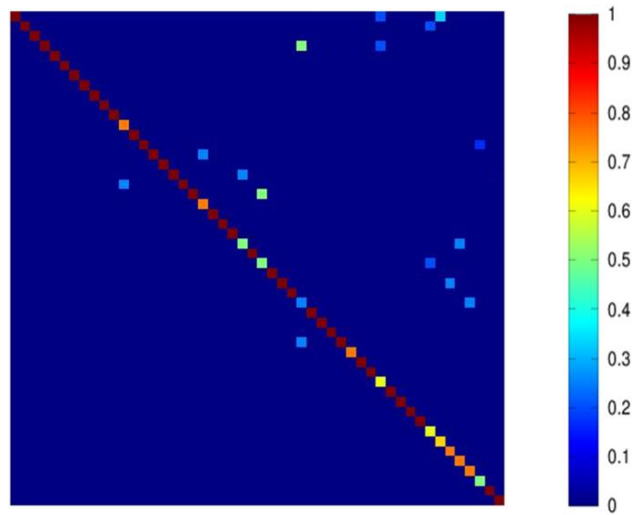

**Fig. 23. Confusion matrix obtained by page-level writer identification (using block-line segmentation approach) on musical sheets.**

### 4.7. Experiment on Musical Sheet after Removing Staff-lines

We have also performed an experiment on writer identification after removing staff-line from the musical document. For this purpose, we considered the dataset of musical sheet images in which the staff-lines were removed. An example of musical sheet without staff-lines is shown in Fig.24. The score-lines were mapped in these images and segmented from for line-level writer identification. Using HMM-based writer identification framework we obtained 67.84% accuracy on line-level. Using line-level score accumulation, the accuracy in page-level has been reached to 90.27%. Note that the features considered are sliding-window based and hence do not capture the component level information as proposed in Gordo et al. (Gordo, Fornes, & Valveny, 2013). With block-line, the performance of page-level accuracy has been reached to 91.54%. Table IV details the results at line and block-line analysis.

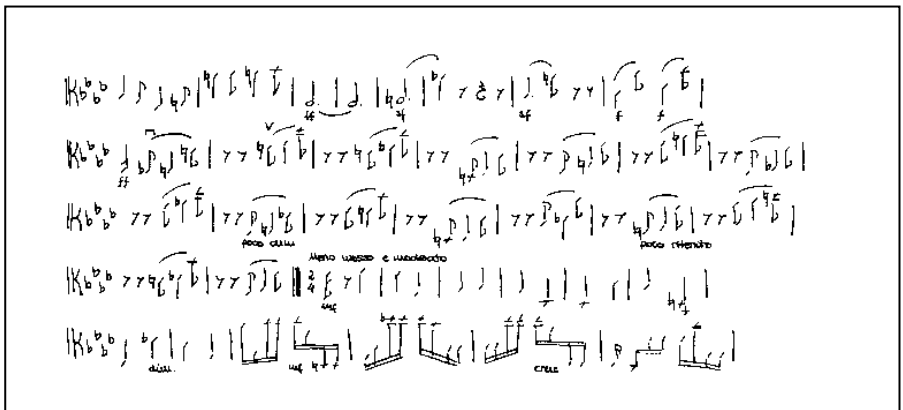

**Fig. 24. Musical sheet image without staff-lines.**

**Table IV: Result of Line level and block-line level approach after staff-line removal**

| Approach | With staff-line | | Without staff-line | |
|---|---|---|---|---|
| Line based performance | Line level | Page level | Line level | Page level |
| | 61.68% | 86.23% | 67.84% | 90.27% |
| Block-line based performance | Block-line level | Page level | Block-line level | Page level |
| | 31.78% | 88.65% | 34.05% | 91.54% |

It was noted that using block-line analysis, the performance of writer identification in page level by removing staff-line was 85.64% without applying FA. By including FA based feature selection, the accuracy increased by 5.90% and we obtained 91.54% overall. Using similar experiment in page level without removing staff-lines the performance increased from 76.98% to 88.65% by applying FA. Here, the performance improved by 11.67% using FA. This improvement shows that the feature selection approach using FA is able to better absorb the noise appearing from staff-lines.

### 4.8. Comparison with other Writer Identification Systems

A comparative study with existing methods CVC-MUSCIMA dataset is given in Table V. Gordo et al. (Gordo, Fornes, & Valveny, 2013) reported 99.7% by using bag of notes features extracted from segmented music-symbols. The best result obtained in ICDAR competition was 77%. These existing approaches assume proper staff-line removal, which may not always be possible due to background noise in documents. Our proposed framework of writer identification without removing staff-lines achieved 88.65% accuracy. Since, the exiting methods removed properly the staff lines manually and hence their results are better than our method (without removing staff lines).

**Table V: Comparison with writer identification approaches.**

| Approach | Removal of staff-line | Feature | Accuracy |
|---|---|---|---|
| Fornes et al. (Fornes, Llados Sanchez & Bunke, 2009) | Yes | Line | 75.00% |
| Fornes et al. (Fornés, Lladós, Sánchez & Bunke, 2008) | | Texture | 73.00% |
| Fornes et al. (Fornés, Lladós, Sánchez, Otazu & Bunke, 2010) | | Line + Texture | 92.00% |
| Fornes et al. (Fornes & Llados, 2010) | | BSM and DTW | 93.00% |
| Gordo et al. (Gordo, Fornes & Valveny, 2013) | | Bag of notes | **99.70%** |
| HMM Framework | No | LGH (Line) | 85.67% |
| | | LGH (Line) +FA | 86.23% |
| | | LGH (Block-line) | 76.98% |
| | | LGH (Block-line) +FA | **88.65%** |
| | | Gabor (Block-line) + FA | 78.45% |

## 4.9. Experiment of Curvy Image

Detection of curvy staff-lines in musical documents is evaluated in this section to understand the effectiveness of our line detection approach. We have applied synthetic curvature generation approach used in Fornes et al. (Kieu, Journet, Visani, Mullot & Domenger, 2013) for evaluation. Sinusoidal curves are generated with musical-symbols for this purpose. Fig.25 shows an example of such curvy image. We applied our HMM-based line detection approach to segment the lines into block-lines and next this block-line level writer-specific information is propagated to page level. Table VI shows the comparative studies of writer identification performance in straight and curve lines. It was noted that, the performance did not drop much in curvy images using block-line analysis. It is due to the fact that segmentation of page was performed in block-line level. In these block-lines, the score-line portions were detected effectively. Hence, the writer-specific scores obtained from block-lines are propagated in page level and we obtained similar writer identification performance as in straight images.

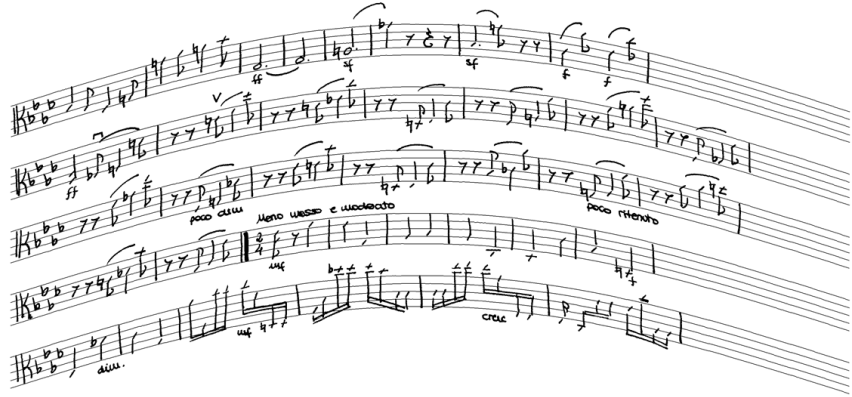

**Fig.25. Example showing synthetically generated curvy image.**

**Table VI: Result of line and block-line segmentation based performance on curvy images**

| Approach | Straight image | | Curved image | |
|---|---|---|---|---|
| Line segmentation based performance | Line level | Page level | Line level | Page level |
| | 61.68% | 86.23% | 43.96% | 75.27% |
| Block-line segmentation based performance | Block level | Page level | Block level | Page level |
| | 31.78% | 88.65% | 30.98% | 86.92% |

## 4.10. Experiment with Synthetic Noise on Musical Scores

We have tested the proposed approach with the musical sheet added with synthetic noises. Different types of noises like Gaussian noise, local noise, 3D noise are added to the original manuscripts to check the robustness of our writer identification approach. Details of these noise generation and writer identification performances are mentioned as follows.

**(a) Gaussian Noise:** The musical sheets are degraded with Gaussian noise of different noise levels (10%, 20% and 30%). Probability density function of Gaussian noise is given by equation (20)

$$PDF_{Gaussian} = \frac{1}{\sqrt{2\pi}\,\sigma} e^{-\frac{(I-\mu)^2}{2\sigma^2}} \dots \dots \dots (20)$$

where, $I$ is is intensity of gray level, $\mu$ and $\sigma$ are mean and standard deviation. To get an idea of such degraded image, a musical sheet image with 30% Gaussian noise is shown in Fig. 26(a). Quantitative results with noisy images obtained by different Gaussian noise levels are shown in Table VII.

**(b) Local Noise:** Fornes et al. (Fornes, Dutta, Gordo & Llados, 2011) applied different local noise generation approaches in CVC-MUSCIMA dataset to mimic old documents' defects. In our experiment, we have considered such noise distortion for writer identification performance. Different types of deformations are added in the dataset to simulate real-world situation: degradation with Kanungo noise, staff-line interruption, typeset emulation, staff-line y-variation, staff-line thickness ratio, staff-line thickness variation, and white speckles. The parameters for each distortion are set according to (Fornes, Dutta, Gordo & Llados, 2011). Fig. 26(c) shows some examples of musical documents after adding such synthetic noise. Quantitative results are shown in Table VII. It was noted that the performance using "staff-line thickness ratio" dropped significantly compared to other approaches.

**(c) 3D Distortions:** 3D distortion model (Kieu, Journet, Visani, Mullot & Domenger, 2013) which was based on 3D meshes and texture coordinate generation was introduced to generate more realistic distortions of the staff lines. Such distortion wrap 2D document image on a 3D mesh acquired by scanning a non-flat old document using a 3D scanner. An example of such musical document after adding 3D distortion is shown in Fig. 26(b). After applying 3D distortion we measured the writer identification performance using our system. Results are detailed in Table VII. We have obtained 80.01% accuracy with block-level performance.

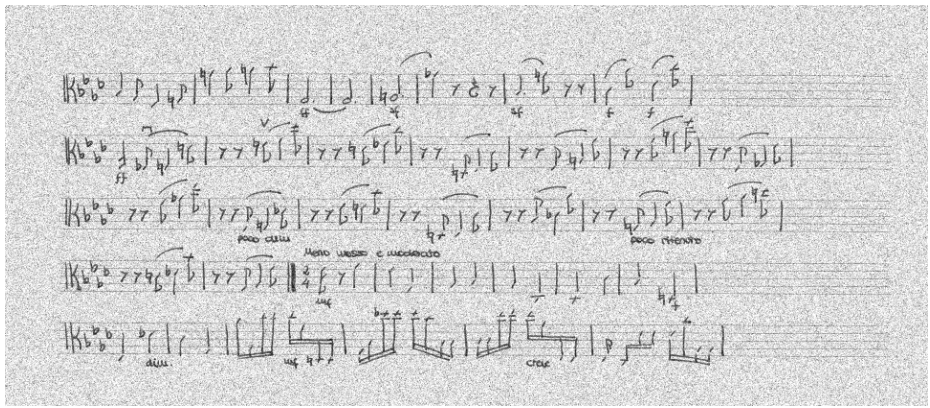

(a)

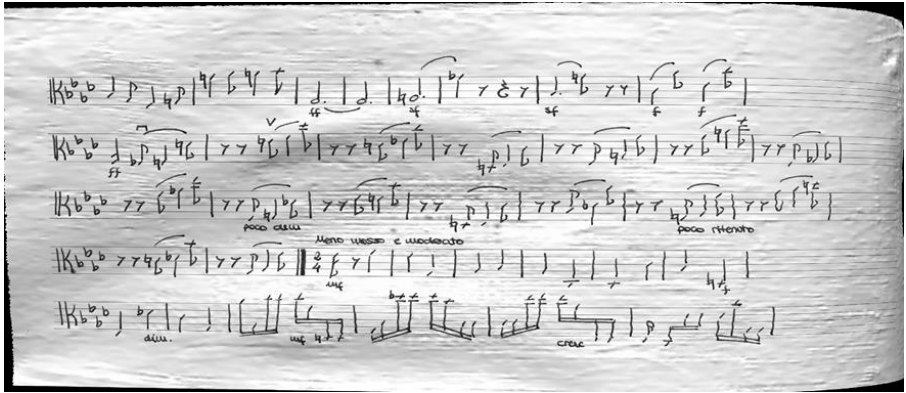

**(b)**

|  |  |  |  |
|---|---|---|---|
| **Ideal Image** |  | **Typeset Emulsion** |  |
| **Kanungo Noise** |  | **Staff-line thickness variation** |  |
| **Interruption** |  | **Staff-line Y variation** |  |
| **White speckles** |  | **Thickness Ratio** |  |

**(c)**

**Fig. 26. Musical sheet with (a) addition of 30% synthetic noise (b) 3D noise and (c) Local noise.**

**Table VII: Writer identification performance in images with Gaussian noise.**

| Deformation Type | | Line based performance | | Block-line based performance | |
|---|---|---|---|---|---|
|  | Noise Parameters | Line level | Page level | Block level | Page level |
| Gaussian Noise | 10% | 61.46% | 86.01% | 31.65% | 88.35% |
|  | 20% | 59.72% | 85.23% | 31.02% | 86.84% |
|  | 30% | 57.52% | 83.16% | 30.12% | 83.87% |
| Local Noise | Typeset Emulation | 61.39% | 85.94% | 31.41% | 88.06% |
| | Kanungo Noise | 60.32% | 84.74% | 30.84% | 87.21% |
| | Staff-line thickness variation | 59.92% | 85.87% | 31.19% | 87.78% |
| | Interruption | 59.94% | 85.47% | 31.09% | 87.59% |
| | Staff-line Y variation | 59.53% | 84.37% | 31.01% | 86.48% |
| | White speckles | 57.33% | 83.37% | 29.64% | 83.38% |
| | Staff-line thickness ratio | 38.21% | 72.34% | 22.66% | 73.18% |
| | 3D Distortion | 55.87% | 80.14% | 29.61% | 81.22% |

## 4.11. Experiment with Historical Musical Documents

Our proposed method could be a better substitute for writer identification system in historical musical documents where the documents are more degraded compared to CVC-MUSCIMA datasets. Most of the staff-line removal methods might fail to remove staff-lines in historical musical documents. This is mainly due to degraded nature and inherent noise of the historical documents. To show the effectiveness of our system, we have collected a total of 218 historical music score images from 34 different writers having 5 to 7 images for each writer[1]. Some example images of this dataset are shown in Fig.27. During this experiment, the parameters were kept fixed according to CVC-MUSCIMA dataset. To measure the performance, we have used 5 fold cross validation, i.e. 4 folds are used as training and rest 1-fold is used as testing. The writer identification accuracy is detailed in Table VIII. The writer identification performance using block-line segmentation was noted as 93.34%. Whereas, we obtained 89.64% accuracy after staff-line removal approach. We can see that the performance without staff-line removal is higher than staff-line removal method. From the result, it is evident that our proposed method performs well in case of historical musical documents. Staff-line removal is not always a good choice where the documents have such inherent noises. Note that, the writer identification performance is little higher in this historical musical dataset compared to CVC-MUSCIMA dataset. In CVC-MUSCIMA dataset, there are twenty different pages and each of these pages is written by all writers. This makes the task of writer identification more difficult in CVC-MUSCIMA dataset. However, it is rare to get multiple copies of same musical script written by different writers in case of historical documents. Hence, we obtained higher accuracy in our historical musical dataset.

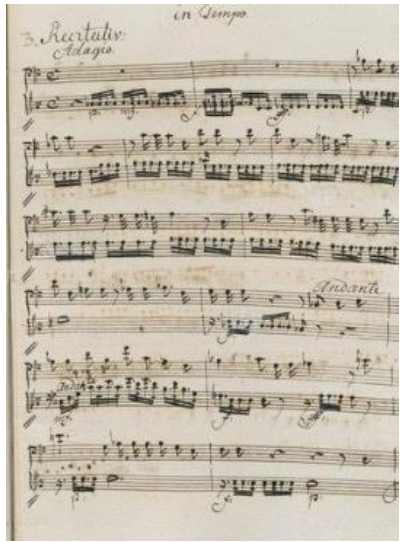 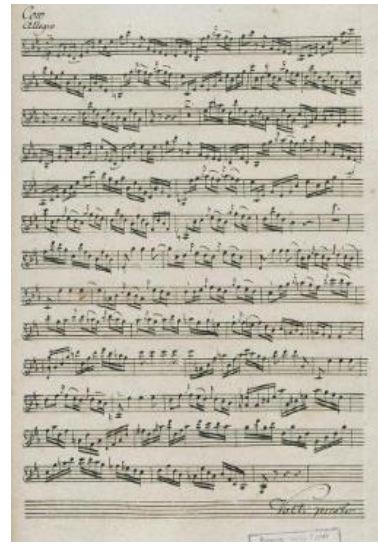

**(a)**                 **(b)**

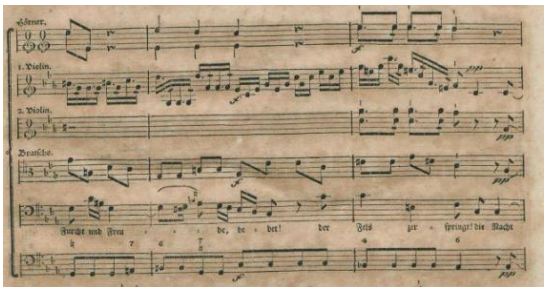 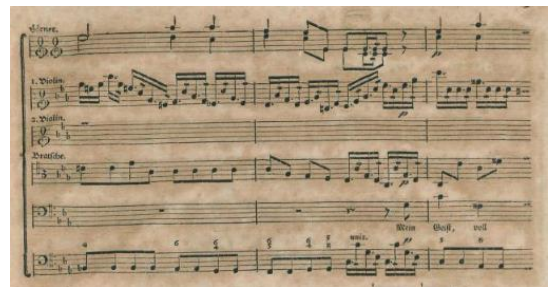

**(c)**                 **(d)**

**Fig.27. Some example images of musical sheets collected from historical musical document archive.**

**Table VIII: Writer identification performance in historical musical documents**

| Line based approach | | Block-line based approach | |
|---|---|---|---|
| Line level | Page level | Block-line level | Page level |
| 64.32 | 90.36 | 42.24 | 93.34 |

---

[1] https://www.bach-digital.de/receive/BachDigitalWork_work_00002259 (accessed on 10th May, 2017)

## 4.12. Runtime evaluation

The proposed framework has been studied using a computer I5 CPU of 2.80 GHz and 4GB RAM 64 bit. MATLAB version 2013b has been used for implementing the system. The performance of execution time has been computed from different runs made in the experiment of CVC-MUSCIMA dataset. The average runtime of writer identification using line level segmentation is 2.12 seconds. Out of it, 0.68 second was consumed for line segmentation and rest 1.44 second was used for feature extraction, line level writer identification and score combination to get the final page level writer identification result. The same for block-line based writer identification are 3.84 seconds, 1.24 seconds and 2.6 seconds respectively. Details of execution time are given in Table VIII. The increased time requirement using block-line based approach was mainly due to more number of segmented parts (termed as block) than score-lines. We also computed the runtime for LGH feature extraction. The average time to extract the LGH feature is 0.12 second. We found an improvement of 0.07 second due to inclusion of silence zone removal in our feature extraction process. Also, the number of time steps (i.e. sliding window position) gets reduced significantly due to silence zone removal, which in turns results in better time efficiency for both training and testing phase of HMM. The runtime can be improved by writing an optimized code to implement this whole framework. Moreover, MATLAB has its own burden of computation, which can be avoided if we use other lower level languages like C, C++, etc. Parallel programming might be another solution which can significantly reduce the overall runtime. The feature and recognition results of different page strip and blocks can be computed in a parallel way which will improve the time efficiency.

**Table VIII: Runtime analysis of writer identification performance**

|  | Line segmentation | Writer identification at each line | Writer identification at page level |
|---|---|---|---|
| **Line based approach** | 0.68 | 1.42 | 0.02 |
|  | **Block-line segmentation** | **Writer identification at each block-line** | **Writer identification at page level** |
| **Block-line based approach** | 1.24 | 2.55 | 0.05 |

# 5. Conclusion

We have presented a novel approach for writer identification in music score documents. In traditional approaches, fair pre-processing needs to be done to remove the staff-lines from a musical score document. However, such pre-processing tasks are challenging and may lose important information which may hurt the writer identification task severely. To avoid this drawback, our proposed system identifies the writer of the musical document without removing staff-lines. For this purpose, an off-the-shelf feature extraction technique, namely LGH is used in HMM based writer classification framework. From the experiment we obtained encouraging results which is competitive with existing literature studies. To the best of our knowledge, this is the first study of writer identification without removing staff-lines. Some of the important contributions of this paper are 1) recognition without staff-line removal, 2) use of silence zone for better recognition, 3) use of block-line segmentation for better accuracy, 4) detection of score and without score zones, and 5) weighted accumulation block/line-level score for page level writer identification. Also, a novel Factor Analysis-based feature selection technique was applied in sliding window feature to reduce the noise appearing from staff-lines which proved efficiency in writer identification performance. The proposed system has been tested on real and synthetic noisy images and we obtained encouraging results. The average runtime performance of writer identification using line level and block level segmentation are 2.12 seconds and 3.84 seconds, respectively. The writer identification accuracy can be improved by applying sophisticated feature or a combination of features along with LGH. Also, different machine learning approaches like SVM, Neural Networks, and Random Forest could be used towards extracting writer-dependent features that may improve the writer identification performance.

## Acknowledgement

The authors thank the anonymous reviewers for their constructive comments and suggestions to improve the quality of the paper.